\documentclass[conference]{IEEEtran}
\IEEEoverridecommandlockouts
\usepackage{cite}
\usepackage{amsmath,amssymb,amsfonts}
\usepackage{algorithmic}
\usepackage{graphicx}
\usepackage{textcomp}
\usepackage{xcolor}
\usepackage[normalem]{ulem}
\usepackage{comment}
\usepackage{amsmath,amsfonts,graphicx,amssymb}
\usepackage{kantlipsum}
\usepackage{bbm}

\newtheorem{lemma}{Lemma}

\newtheorem{theorem}{Theorem}

\usepackage{enumerate}
\usepackage{algorithm, algorithmic}

\usepackage{geometry}
\def\BibTeX{{\rm B\kern-.05em{\sc i\kern-.025em b}\kern-.08em
    T\kern-.1667em\lower.7ex\hbox{E}\kern-.125emX}}
\begin{document}

\title{Fixed-Confidence Best Arm Identification \\ with Decreasing Variance}
\author{%
  \IEEEauthorblockN{Tamojeet Roychowdhury$^1$, Kota Srinivas Reddy$^2$, Krishna P Jagannathan$^2$, and Sharayu Moharir$^1$}
  \IEEEauthorblockA{$^1$Department of Electrical Engineering, IIT Bombay, $^2$Department of Electrical Engineering, IIT Madras \\ Email: tamojeet@iitb.ac.in, ksreddy@ee.iitm.ac.in, krishnaj@ee.iitm.ac.in, sharayum@ee.iitb.ac.in}
}

\maketitle
\begin{abstract}
We focus on the problem of best-arm identification in a stochastic multi-arm bandit with temporally decreasing variances for the arms' rewards. We model arm rewards as Gaussian random variables with fixed means and variances that decrease with time. The cost incurred by the learner is modeled as a weighted sum of the \emph{time needed} by the learner to identify the best arm, and the \emph{number of samples} of arms collected by the learner before termination. Under this cost function, there is an incentive for the learner to not sample arms in all rounds, especially in the initial rounds. On the other hand, not sampling increases the termination time of the learner, which also increases cost. This trade-off necessitates new sampling strategies. We propose two policies. The first policy has an initial wait period with no sampling followed by continuous sampling. The second policy samples periodically and uses a weighted average of the rewards observed to identify the best arm. We provide analytical guarantees on the performance of both policies and supplement our theoretical results with simulations which show that our polices outperform the state-of-the-art policies for the classical best arm identification problem. 
\end{abstract}
\color{black}

\section{Introduction}
Consider the setting where a large pool of reviewers is given competing products $K$ to use their feedback to identify the best product. All reviewers start using the products at the same time, and with each subsequent use, the accuracy of each reviewer's estimate of the product's utility increases. Thus, the reviewers' feedback becomes more reliable over time. The goal is to identify the product with the highest utility as quickly as possible by querying as few reviewers as possible.

Motivated by this, we study the task of best arm identification in the multi-arm bandit setting where the rewards of arms are nonstationary. Specifically, arm rewards are Gaussian with time-invariant means, but their variance decreases over time. The total cost incurred by the learner is modeled as the sum of the number of rounds taken by the learner to identify the best arm and the cumulative sampling cost incurred up to that round, where the sampling cost incurred in a round is proportional to the number of arms sampled in that round. Our goal is to identify the best arm, defined as the arm with the highest mean reward, while minimizing the cost incurred by the system, subject to an upper bound on the probability of identifying a wrong arm as the best arm. This is the widely studied \emph{fixed confidence} setting for the best arm identification problem in multi-arm bandits \cite{garivier2016optimal, jamieson2014best,even2006action}.

\subsection{Analytical Challenges and Contributions}
In the classical version of best arm identification in multi-arm bandits, the reward distributions of the arms are time-invariant. The cost incurred by the learner is the number of samples needed by the learner to identify the best arm \cite{garivier2016optimal,jamieson2014best}. In our problem, since the variance of the arms decreases over time and the cost incurred by the learner increases with the number of arms sampled, unlike the classical version of the problem, there is an incentive to not sample arms in all rounds, especially, in the initial rounds when the variance in the rewards observed is high. However, since the cost incurred is also an increasing function of the time needed to identify the best arm, there is a trade-off between reducing sampling cost and minimizing the number of rounds needed to identify the best arm. This necessitates new sampling strategies. The key contributions of this work are as follows:
\begin{enumerate}
 \item[(a)] We propose a sampling policy for the setting where the gap between the means of the best and the second best arm is known and provide performance guarantees. 

 \item[(b)] We propose a policy for the setting in which no side information is known about the means of the $K$ arms. The key novelty in this policy is twofold: \emph{(i)} periodic sampling, i.e. arms are sampled periodically and the period is an increasing function of the number of arms and the cost of sampling per arm, \emph{(ii)} the empirical estimate of the reward of an arm is a weighted average of the reward samples collected where the weights increase with the round in which each sample has been collected. The motivation behind this is that the samples collected in subsequent rounds have lower variance and therefore are more representative of the mean rewards of the various arms. We provide performance guarantees for the proposed policy. 
 
 \item[(c)] Through simulations, we show that our policies outperform the state-of-the-art policies for the classical problem of best-arm identification in multiarm bandits.
\end{enumerate}

\subsection{Related Works}
While our focus is on the identification of the best arm, another widely studied problem in multiarm bandits is regret minimization \cite{lattimore2020bandit,auer2010ucb}. The regret minimization problem has been studied for many variants of the multiarm bandits with nonstationary rewards. The seminal work on nonstationary bandits is the setting of restless bandits \cite{whittle1988restless} in which the state of the arms and therefore their reward distribution change in a Markovian manner. Multiple other variants of the problem of regret minimization in multiarm bandits with non-stationary rewards have been studied \cite{garivier2008upper, krishnamurthy2021slowly, auer2019adaptively, besbes2014stochastic, wei2018abruptly,bacchiocchi2024autoregressive}. In \cite{garivier2008upper}, the focus is on the setting in which the reward distributions remain constant over epochs and change at unknown time instants. In \cite{krishnamurthy2021slowly}, the authors consider the setting in which the arms' rewards are stochastic and independent over time, but the absolute difference between the expected rewards of any arm at any two consecutive rounds is bounded by a drift limit. The authors propose a policy and provide an instance-dependent regret upper bound for it. In \cite{auer2019adaptively}, the authors consider a multiarm bandit with stochastic reward
distributions that change abruptly several times and achieve (nearly) optimal mini-max regret bounds without knowing the number of changes. In \cite{besbes2014stochastic}, the focus is on a non-stationary multiarm bandit setting, and the authors show the connection between the extent
of allowable reward ``variation" and the minimal achievable regret. In \cite{wei2018abruptly}, the focus is on abruptly changing and slowly varying environments. The authors propose variants of the UCB algorithm with sublinear regret for their setting. In \cite{bacchiocchi2024autoregressive}, the authors consider regret minimization where the reward process follows an autoregressive model.

The best arm identification problem in non-stationary environments has received relatively limited attention. For a survey on  best arm identification with fixed confidence, see \cite{jamieson2014best}. In \cite{allesiardo2017non}, the authors study a generalization of the stationary stochastic bandit problem in which the adversary chooses before a sequence of distributions that govern the award of an arm over time. The authors define an appropriate notion of sample complexity for the setting where the best arm changes over time, propose a variant of successive elimination, and provide performance guarantees for it. In \cite{ghatak2024best}, the focus is on the best arm identification problem in nonstationary environments motivated by the problem of beam selection in mm-wave communication. To the best of our knowledge, the problem we focus on in this work has not been studied in the existing literature. 
\color{black}

\section{Problem Setup and Preliminaries}
We consider a multi-armed bandit setting with $K$ arms, where each arm generates rewards according to a Gaussian distribution with an unknown mean and a known, time-varying variance. In particular, we assume that the reward of arm $K$ in round $t$ is distributed as $ \mathcal{N}(\mu_k, \sigma^2 / t) $. The problem instance, denoted by $\nu$, is uniquely defined by the $K$-dimensional vector $ \nu = [\mu_1, \mu_2, \dots, \mu_K] $. 

The best arm, i.e., arm $k^*$, where $$ k^* = \arg\max_{1\leq k \leq K} \mu_k,$$ is assumed to be unique. Without loss of generality, we assume that arms are indexed in decreasing order of their means, i.e., $\mu_i \geq \mu_j$ for $i \leq j$. It follows that $k^* = 1$. We allow for multiple arms to be sampled in a round. Let $\mathcal{A}_t$ denote the set of arms sampled in round $t$ and let $\mathcal{X}_t$ denote the set of corresponding rewards observed in round $t$. Let $\mathcal{A}_{1:t}$ denote the sequences of sets of arms sampled in rounds $1$ to $t$ and $\mathcal{X}_{1:t}$ denote the sequences of sets of the corresponding rewards observed in rounds $1$ to $t$. A policy $\pi = \{\pi_t\}_{t=1}^{\infty}$ is a mapping from ($\mathcal{A}_{1:t}$, $\mathcal{X}_{1:t}$) to one of two possible decisions: stop and declare the estimated best arm or select $\mathcal{A}_{t+1}$. Let $\tau_{\pi}$ denote the (possibly random) stopping time under the policy $\pi$ and $\eta_{\pi}$ denote the cumulative number of arms sampled in rounds 1 to $\tau_{\pi}$ under the policy $\pi$. The cost incurred by the policy $\pi$, denoted by $\mathcal{C}_{\pi}$, is defined as:
\begin{align}
\label{eq:cost}
\mathcal{C}_{\pi} = \tau_{\pi} + c \eta_{\pi},
\end{align}
where $c>0$ is the cost incurred to obtain one sample of one arm. For a given $\delta \in (0,1)$, $\Pi(\delta)$ is the class of policies for which the estimated best arm is $k^*$ with probability at least $1-\delta$. Our goal is to design a low-cost policy belonging to $\Pi(\delta)$.

Let $\tilde{\mu} = \max_{1 \leq k \leq K, k \ne k^*} \mu_k = \mu_2$. Then, the sub-optimality gap, denoted by $\Delta$ is defined as:
\begin{align}
\label{eq:gap}
    \Delta = \mu_{k^*} - \tilde{\mu} = \mu_1 - \mu_2.
\end{align}
Further, for $k\geq 3$, we define 
\begin{align}
\label{eq:gap_general}
\Delta_k = \mu_1 - \mu_k. 
\end{align}

\section{Algorithms and Performance Guarantees}
In this section, we present our algorithms and provide guarantees on their performance. We consider two settings: the first where the difference between the means of the best and the second best arms is known and the second setting where the learner has no side information about the means of the $K$ arms. 
\subsection{Known Sub-optimality Gap}
We first focus on the setting where the sub-optimality gap $\Delta$ is known.  
Our algorithm for this setting is designed on the principle that if we are to collect a fixed number of samples of each arm before we identify the best arm, the optimal time to collect those samples would be the block of rounds right before we output the estimated best arm. This is a consequence of the fact that the variance of samples is lower in the later rounds. Motivated by this, we propose an algorithm called Wait Then Continuously Sample (WTCS) which operates in two phases. In Phase 1, the algorithm waits, i.e., it does not sample any arms. In Phase 2, the algorithm samples all arms in each round. At the end of Phase 2, the algorithm outputs the arm with the highest empirical mean as the estimated best arm. It follows that the duration of the two phases completely characterizes the algorithm.

Let $\Delta$ denote the difference between the means of the best as second best arms as defined in \eqref{eq:gap} and let 
\begin{align}
    t_W = \frac{2\sigma}{\Delta} \sqrt{Kc\cdot \ln \left( \frac{K}{\delta} \right)}.
\end{align}
The WTCS policy samples all arms in rounds $t_W+1$ to $t_W + \frac{t_W}{Kc}$ and outputs the arm with the highest empirical mean at the end of round $t_W + \frac{t_W}{Kc}$ as the estimated best arm. 
Next, we characterize the performance of WTCS.
\begin{theorem}
\label{thm:WTCS}
   For a given $\delta \in (0,1)$, WTCS $\in \Pi(\delta)$ with
\begin{align*}
    \mathcal{C}_{\text{WTCS}}  = \frac{6\sigma}{\Delta} \sqrt{Kc \cdot \log \left( \frac{K}{\delta} \right)} .
\end{align*}
\end{theorem}

\subsection{No Side-information about Arm Means}
 The WTCS policy discussed above requires the knowledge of the sub-optimality gap $(\Delta)$ and uses it to determine how long to wait before sampling arms. While an initial wait period with no sampling is the right approach for our problem, without the knowledge of the optimality gap, it is difficult to determine how long to wait. Waiting for too long increases the time needed to output the estimated best arm while not waiting enough leads to an increase in the sampling cost as the early samples tend to be noisy.

We propose a new policy called Periodic Sampling with Weighted Successive Elimination (PS-WSE). PS-WSE  balances the trade-off between the sampling cost and the cost incurred due to the time taken to identify the best arm by sampling periodically at a frequency that decreases with the sampling cost and the number of arms. 
\color{black}

PS-WSE maintains a set of active arms and samples all active arms periodically. The period of sampling, denoted by $\lambda,$ is fixed to $cK$. 
In rounds in which the active arms are sampled, i.e., for $t$ such that $t \mod \lambda = 0$, we compute an empirical estimate of the reward of each arm. This empirical estimate is a weighted average of the rewards observed up to that time with higher weights given to samples collected in later rounds. This is motivated by the fact that the variance of the arm reward decreases with time and therefore, samples collected later are more representative of the mean rewards as compared to earlier samples. Let $X_{j,t}$ denote the reward for arm $j$ in round $t$. Then, the empirical mean of the reward of arm $j$ at the end of sampling round $r=t/\lambda$, denoted by $\hat{\mu}_j(r)$ is given by
\begin{align}
\label{eq:weighted_mean}
    \hat{\mu}_j(r) = \sum_{\tau = 1}^{r} w_{\tau, r} X_{j,\tau\lambda}, \text{ where, } w_{\tau, r} = \frac{2\tau}{r(r+1)}.
\end{align}
Following this, we eliminate arms that have empirical reward estimates below a carefully chosen time-varying threshold from the active set. The last remaining arm in the active set is declared as the best arm by the PS-WSE policy. 

Refer to Algorithm \ref{alg:cap} for a formal definition of PS-WSE. 
\begin{algorithm}
\caption{Periodic Sampling with Weighted Successive Elimination (PS-WSE)}\label{alg:cap}
\begin{algorithmic}[1]
\STATE Input: $c$, $K$, $\sigma$, $\delta$
\STATE $S \gets \{1, 2, \cdots, K\}$, $\hat \mu_1, \hat \mu_2 \ldots \hat \mu_K \gets 0$, $\lambda \gets  cK$
\WHILE{$t > 0$}
\IF{$|S|>1$ and $t \mod \lambda = 0$}
\STATE $r=\frac{t}{\lambda}$
\FOR{$j \in S$} 
\STATE Sample arm $j$ to obtain $X_{j,t}$
\STATE Update $\hat{\mu}_j$ according to \eqref{eq:weighted_mean}
\ENDFOR
\STATE $\hat{i}^* \gets \arg\max_{i \in S} \hat{\mu}_i$
\STATE $S \gets S \setminus \left\{ j: \hat{\mu}_j < \hat{\mu}_{\hat{i}^*} - \frac{2\sigma}{t} \sqrt{\lambda \cdot \log \left( \frac{2Kt^2}{\lambda^2\delta} \right)} \right\}$
\ENDIF
\IF{$|S| = 1$}
\STATE Output the arm in $S$ as the optimal arm
\STATE Break
\ENDIF
\ENDWHILE
\end{algorithmic}
\end{algorithm}

We now characterize the performance of PS-WSE. 
\begin{theorem}
\label{thm:PS-WSC}
   For a given $\delta \in (0,1)$, PS-WSE $\in \Pi(\delta)$, and in the event where PS-WSE identifies the best arm correctly,    
\begin{align*}
   \mathcal{C}_{\text{PS-WSE}}  \leq & \frac{6\sigma(cK+2c)}{\Delta} \sqrt{ \frac{1}{cK} \ln \left( \frac{36  \sigma^2}{c\delta\Delta^2} \right)} \\ & \quad \quad + \sum_{j=3}^K \frac{6c\sigma}{\Delta_j} \sqrt{ \frac{1}{cK} \ln \left( \frac{36\sigma^2}{c\delta\Delta_j^2} \right)} \\
     =& \mathcal{O}\left( \frac{\sigma}{\Delta}  \sqrt{Kc \log \left(\frac{36\sigma^2}{c \delta  \Delta^2}\right)} \right).
\end{align*}

\end{theorem}

\section{Simulation Results}
In this section, we compare the performance of our policies with that of the state-of-the-art policies for the fixed confidence setting for the classical best arm identification problem where rewards of arms are stationary. In addition to WTCS and PS-WSE, we simulate the following policies:
\begin{enumerate}
    \item[1.] \emph{Successive Elimination} (SE) \cite{even2006action, jamieson2014best}: A set of active arms is maintained. In each round, all active arms are sampled. All arms whose UCB is lower than the LCB of the arm with the highest empirical reward are eliminated from the active set. The estimated best arm is the last arm left in the active set. 
    \item[2.] \emph{Lower Upper Confidence Bound} (LUCB) \cite{kalyanakrishnan2012pac, jamieson2014best}: Each arm is initially sampled once. Following this, in each round, we sample the arm with the highest empirical mean and the arm with the highest UCB (excluding the arm with the highest empirical mean). We stop when there exists an arm whose LCB is higher than the UCB of all other arms and this arm is identified as the best arm. 

\end{enumerate}

\subsection{Simulation Settings}
\begin{figure*}[h!]
	\includegraphics[width=0.9\columnwidth]{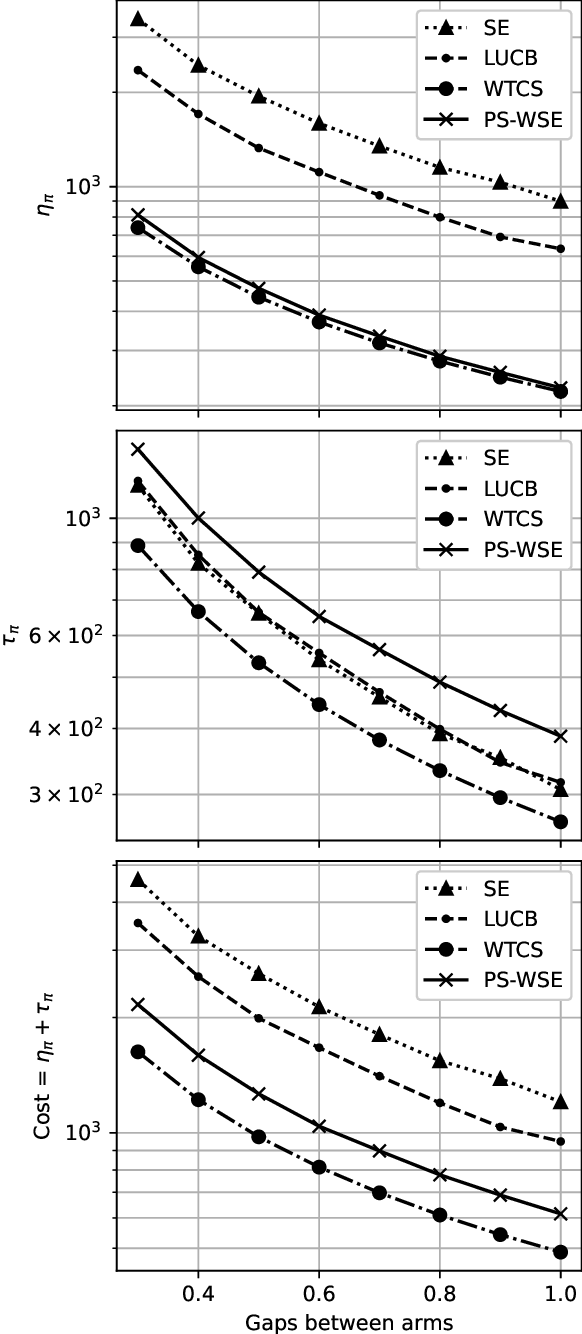}\ \ \ \ \ \ 
    \includegraphics[width=0.84\columnwidth]{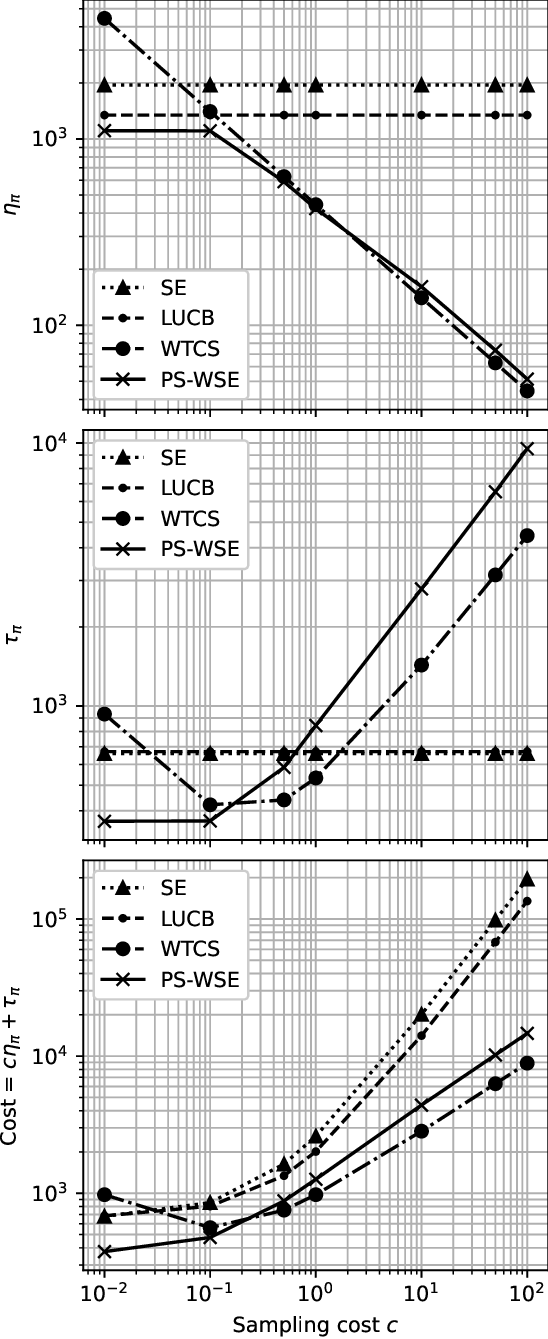}
    \caption{The three plots on the left are for Experiment 1 (varying sub-optimality gap) and the ones the right are for Experiment 4 (varying sampling cost $c$). We plot the number of samples, the stopping time, and the total cost incurred by all four candidate policies.}
    \label{fig:sim-results-1}
\end{figure*}
\noindent We conduct the following four sets of experiments. 
\begin{enumerate}
\item[1.] The number of arms is fixed to 5, and the arm means are in an arithmetic progression with the common difference being a varying parameter, from 0.3 to 1 in steps of 0.1. The cost of sampling an arm, $c$, is fixed to 1 and $\sigma = 10$. 

\item[2.] The arm means are equally spaced in the interval $[0,3]$, $c=1$, and $\sigma=10$. We vary the number of arms from 2 to 12 in steps of 2. 

\item[3.]
The number of arms is fixed to 5, with any two consecutive arms differing in their means by 0.5, $c=1$.  We vary $\sigma$ from 1 to 11 in steps of 2.

\item[4.]
The number of arms is fixed to 5, with any two consecutive arms differing in their means by 0.5, $\sigma$ is fixed at 10, and $c$ is varied from 0.01 to 100. 
\end{enumerate}

\subsection{Results}

\begin{figure*}[h]
    \includegraphics[width=0.85\columnwidth]{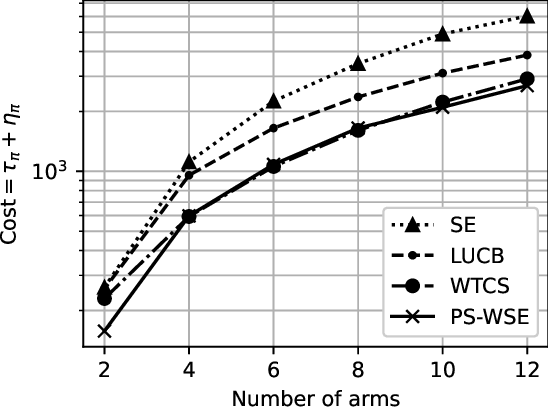}\ \ \ \ \ \
    \includegraphics[width=0.85\columnwidth]{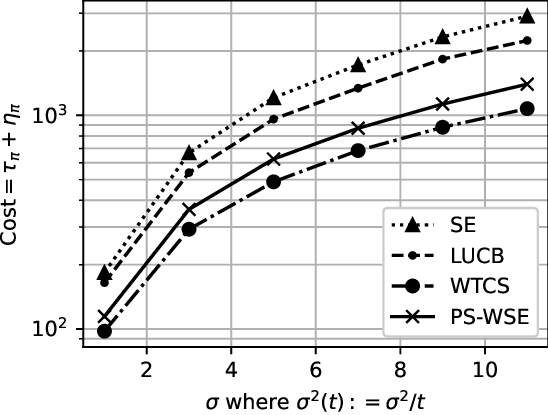}
    \caption{Cost as a function of the varying parameter for varying numbers of arms on the left and for varying $\sigma$ on the right for all four candidate policies.}
    \label{fig:sim-results-2}
\end{figure*}

The individual cost components, $\tau_\pi$ and $\eta_\pi$ are shown for Experiment 1 (varying arm gaps) and Experiment 4 (varying sampling cost $c$) are shown in Fig~\ref{fig:sim-results-1}. Due to the time interval in periodic sampling, the stopping time $\tau_\pi$ is larger for PS-WSE than for all other policies. This, however, is more than compensated by the large reduction in the number of samples $\eta_\pi$, giving a lower overall cost than both SE and LUCB. Similar trends are observed in the individual cost components for Experiments 2 and 3 as well. For Experiment 4, as the sampling cost $c$ increases, we see a drop in the number of samples and a rise in stopping time for both WTCS and PS-WSE, as expected. For WTCS the number of samples $\eta_\pi$ varies inversely as $\sqrt{c}$, while the stopping time $\tau_\pi \sim (1+\frac{1}{c})\sqrt{c}$ which can be seen in the minima. PS-WSE behaves similarly, except for very low $c$ where the sampling period $\lambda = Kc<1$, is rounded off to 1 and PS-WSE results in continuous sampling, making it similar to SE. The weighing of samples however still gives it an overall lower cost.

The overall cost for Experiments 2 and 3 are shown in Fig~\ref{fig:sim-results-2}. We note that in all four experiments, our algorithms perform significantly better compared to the standard algorithms for the multi-arm bandit problem.

\section{Proofs}

In this section, we discuss the outlines of the proofs of Theorems \ref{thm:WTCS} and \ref{thm:PS-WSC}.

\subsection{Proof Outline for Theorem \ref{thm:WTCS}}

Recall that the WTCS policy samples all arms in rounds $ t_W + 1 $ to $ t_W + \frac{t_W}{Kc} $ and outputs the arm with the highest empirical mean at the end of round $ t_W + \frac{t_W}{Kc}  $ as the estimated best arm. Thus, the total number of samples is $ \eta_{\pi} = K \left( \frac{t_W}{Kc}\right) = \frac{t_W}{c} $, and the time of the last sample is $ \tau_{\pi} = t_W + \frac{t_W}{Kc} $. Therefore, the cost is given by
\begin{align*}
   \tau_{\pi} + c \eta_{\pi} &= t_W + \frac{t_W}{Kc} + c \left( \frac{t_W}{c}  \right) \nonumber \\
   &= \frac{2 \sigma}{\Delta} \left( 2 + \frac{1}{Kc} \right) \sqrt{Kc \cdot \ln \left( \frac{K}{\delta} \right)} .
\end{align*}

Now, we need to prove that the fixed-confidence condition is satisfied. Note that by our algorithm, an error occurs only if, at the end of sampling, one of the arms $ j \neq 1 $ has a higher empirical mean than arm 1. Let $ \hat{\mu}'_j $ denote the empirical mean of arm $ j $ at the end of sampling by the WTCS algorithm. Then,
$$\hat{\mu}'_j = \sum_{t=t_W+1}^{t_W + \frac{t_W}{Kc}} \frac{X_{j,t}}{\left(\frac{t_W}{Kc} \right)}.$$

The event that an arm $ j \neq 1 $ is declared as the best arm by the WTCS algorithm implies that $ \hat{\mu}'_j \geq \hat{\mu}'_1 $. We can verify that the probability
$$\mathbb{P}(\hat{\mu}'_j \geq \hat{\mu}'_1) \leq \frac{\delta}{K}.$$
Therefore, by applying the union bound, we conclude that the WTCS algorithm identifies an incorrect arm as the best arm with probability less than $ \delta $.

\subsection{Proof Outline for Theorem \ref{thm:PS-WSC}}

The proof of Theorem \ref{thm:PS-WSC} follows along the lines of the proofs
of \cite[Theorems 1 and 2]{srinivas2022almost} and uses the following lemma.
 
\begin{lemma}
At a sampling time $ t $, let $ U_t \triangleq \frac{\sigma}{t} \sqrt{\lambda \cdot \ln{ \left( \frac{2K t^2 }{ \lambda^2 \delta} \right)} } $ and let $ \mathcal{E} $ be an event defined as follows:
\begin{equation*}
	\mathcal{E} := \bigcap_{t \in \mathbb{N}, k \in [K]} \left\{ \left| \hat{\mu}_j(t) -  \mu_j \right| \leq {U_t} \right\}.
\end{equation*}
Then, $ \mathbb{P}(\mathcal{E}) \geq 1 - \delta. $
\end{lemma}
First, we prove that the PS-WSE algorithm is a $\Pi(\delta)$ algorithm using the proof by contradiction technique.  
Suppose the algorithm declares an arm \(j \neq 1\) as the best arm. According to the algorithm, there exists a sampling time \(t\) such that  
\begin{align}\label{eqn:aaa}
    \hat{\mu}_j(t) - \hat{\mu}_1(t) > 2U_t.
\end{align}  
However, under the event \(\mathcal{E}\), for every sampling time \(t\), we have  
\begin{align}\label{eqn:bbb}
    \hat{\mu}_1(t) - \hat{\mu}_j(t) \leq (\mu_1 + U_t) - (\mu_j - U_t) = 2U_t.
\end{align}  
Equations \eqref{eqn:aaa} and \eqref{eqn:bbb} are contradictory. Thus, under the event \(\mathcal{E}\), the PS-WSE algorithm always identifies the correct arm. Therefore, the PS-WSE algorithm is a $\Pi(\delta)$ algorithm.

Next, we focus on the cost of the PS-WSE algorithm. Under the event $\mathcal{E}$,
\begin{align*}
    \hat \mu_{j} &\in \left( \mu_j - U_t, \mu_j + U_t \right). 
\end{align*}

Since the algorithm always outputs the correct best arm under the event $ \mathcal{E} $, any other arm $ j $ is eliminated when $ \hat{\mu}_{1} - \hat{\mu}_{j} > 2U_t $. We can show that for $ t = T_j = \frac{6 \sigma}{\Delta_j} \sqrt{\lambda \ln \left( \frac{36 K \sigma^2}{\lambda \delta \Delta_j^2} \right)} $, the condition $ \hat{\mu}_{1} - \hat{\mu}_{j} > 2U_t $ will be satisfied. Therefore, arm $ j $ is surely eliminated by the end of round $ T_j $. Since $ T_2 $ is the highest value, we also have $ \tau_\pi \leq T_2 $. Further, in the scenario where all sub-optimal arms are at the same gap from the best arm $(\Delta_j=\Delta\ \forall j)$, the cost $\mathcal{C}_{\rm{PS-WSE}} \propto \left( \sqrt{\lambda} + \frac{cK}{\sqrt{\lambda}} \right)\sqrt{\ln\left( \frac{1}{\lambda} \right)}$. Ignoring the root logarithmic term, the minimum is achieved at $\lambda=cK$. Therefore, using this,
\begin{align*}
    \mathcal{C}_{\rm{PS-WSE}} =& \tau_\pi + c\eta_\pi \leq T_2 + 2c\left(\frac{T_2}{Kc} \right) + c \sum_{j=3}^K \frac{T_j}{Kc} \\
    =& \frac{6\sigma(cK+2c)}{\Delta} \sqrt{ \frac{1}{cK} \ln \left( \frac{36  \sigma^2}{c\delta\Delta_j^2} \right)} \\ & \hspace{0.75in}+ \sum_{j=3}^K \frac{6c\sigma}{\Delta_j} \sqrt{ \frac{1}{cK} \ln \left( \frac{36 \sigma^2}{c\delta\Delta_j^2} \right)}.
\end{align*}

\section{Conclusions and Future Work}
We focus on the problem of best arm identification in a stochastic multi-arm bandit setting where arm rewards have fixed means and temporally decreasing variances. The cost is modeled as a linear combination of the number of rounds needed to identify the best arm and the number of reward samples collected before declaring the estimated best arm. 

We propose two sampling policies and provide performance guarantees for them. The first policy requires the knowledge of the sub-optimality gap and uses it to determine an initial wait period when no arms are sampled. Following this, the policy samples all arms for a fixed duration and outputs the arm with the highest empirical mean at the end of this period as the best arm. The second policy samples arms periodically at a frequency that decreases with the number of arms and the sampling cost. It computes an appropriately weighted average of the samples collected from each arm and uses it to estimate the best arm. Via simulations, we show that our policies outperform popular policies for the classical best arm identification problem. 

Characterizing a fundamental lower bound on the expected cost incurred by any online policy for our setting remains an open problem. Further, an interesting variant of our problem is the setting where in addition to decreasing variances, samples are correlated across time \cite{gupta2021best}.

\section*{Acknowledgements}
Kota Srinivas Reddy's work is supported by the Department of Science and Technology (DST), Govt. of India, through the INSPIRE faculty fellowship. Sharayu Moharir's work is supported by a SERB grant on Leveraging Edge Resources for Service Hosting.

\color{black}

\bibliographystyle{IEEEtran}
\bibliography{IEEEabrv,references}

\end{document}